\newif\ifarxiv
\newif\ifralfinal
\newif\ifconffinal
\ralfinalfalse \arxivtrue \conffinalfalse

\ifarxiv
\documentclass[letterpaper, 10 pt, conference]{ieeeconf}  %
\fi
\ifralfinal
\documentclass[letterpaper, 10 pt, journal, twoside]{IEEEtran}
\fi
\ifconffinal
\documentclass[letterpaper, 10 pt, conference]{ieeeconf} %
\fi

\IEEEoverridecommandlockouts

\usepackage{graphics} %
\usepackage{epsfig} %
\usepackage{times} %
\usepackage{amsmath} %
\usepackage{amssymb}  %
\usepackage{bbm}
\usepackage{booktabs}
\usepackage{mathtools}
\usepackage{multirow}
\usepackage{units}
\usepackage{bm}
\usepackage{xcolor}
\let\labelindent\relax
\usepackage{enumitem}
\usepackage{balance}
\usepackage[ruled,vlined]{algorithm2e}
\usepackage{algorithmicx}
\usepackage{cite} 
\usepackage[caption=false, font=footnotesize]{subfig}
\usepackage{graphicx}
\usepackage{adjustbox}
\usepackage{letltxmacro}
\LetLtxMacro{\originaleqref}{\eqref}
\renewcommand{\eqref}{Eq.~\originaleqref}
\usepackage{algpseudocode}
\algnewcommand\Algand{\textbf{and} }
\newcommand{\eg}{\emph{e.g.},}

\SetCommentSty{mycommfont} %
\usepackage{microtype}

\definecolor{LightCyan}{rgb}{0.88,1,0.88}
\definecolor{emb_color}{RGB}{252,224,225}
\definecolor{multi_head_attention_color}{RGB}{252,226,187}
\definecolor{add_norm_color}{RGB}{242,243,193}
\definecolor{ff_color}{RGB}{194,232,247}
\definecolor{softmax_color}{RGB}{203,231,207}
\definecolor{linear_color}{RGB}{220,223,240}
\definecolor{gray_bbox_color}{RGB}{243,243,244}

\ifarxiv
\usepackage{fancyhdr}
\fi
\makeatletter
\let\NAT@parse\undefined
\makeatother
\usepackage{hyperref}  %

\DeclareMathOperator*{\argmin}{arg\,min}

\newcommand{\algoname}{Reg-NF}
\newcommand{\dataname}{ONR }
\begin{document}

\title{
\LARGE \bf

Reg-NF: Efficient Registration of Implicit Surfaces within Neural Fields
}
\author{Stephen Hausler$^{1, \dagger}$, David Hall$^{1,\dagger}$, Sutharsan Mahendren$^{1,2}$, Peyman Moghadam$^{1,2}$   
\thanks{$^\dagger$ Equal Contribution Website: \href{https://csiro-robotics.github.io/Reg-NF/}{https://csiro-robotics.github.io/Reg-NF}}
\thanks{$^1$ Authors are with the CSIRO Robotics, DATA61, CSIRO, Brisbane, QLD 4069, Australia. 
E-mails: {\tt\footnotesize \emph{firstname.lastname}@csiro.au}}
\thanks{
$^{2}$  Sutharsan Mahendren, and Peyman Moghadam are with the SAIVT research programme in the School of Electrical Engineering and Robotics, Queensland University of Technology (QUT), Brisbane, Australia.
E-mails: {\tt\footnotesize \emph\{sutharsan.mahendren, peyman.moghadam\}@qut.edu.au}
}
} 

\bstctlcite{IEEEexample:BSTcontrol}

\maketitle
\ifarxiv
\thispagestyle{fancy}
\pagestyle{plain}
\fi

\begin{abstract}

Neural fields, coordinate-based neural networks, have recently gained popularity for implicitly representing a scene. In contrast to classical methods that are based on explicit representations such as point clouds, neural fields provide a continuous scene representation able to represent 3D geometry and appearance in a way which is compact and ideal for robotics applications. However, limited prior methods have investigated registering multiple neural fields by directly utilising these continuous implicit representations. In this paper, we present Reg-NF, a neural fields-based registration that optimises for the relative 6-DoF transformation between two arbitrary neural fields, even if those two fields have different scale factors. Key components of Reg-NF include a bidirectional registration loss, multi-view surface sampling, and utilisation of volumetric signed distance functions (SDFs). We showcase our approach on a new neural field dataset for evaluating registration problems. We provide an exhaustive set of experiments and ablation studies to identify the performance of our approach, while also discussing limitations to provide future direction to the research community on open challenges in utilizing neural fields in unconstrained environments.

\end{abstract}

\section{Introduction}

For robotics applications, the six degree of freedom (6-DoF) registration between two scenes of interest is a crucial step, for tasks such as localisation, object pose estimation and 3D reconstruction. 
While many methods exist for representing 3D scenes, including point clouds, voxels and meshes, recently \emph{implicit representations} have emerged, which can compactly represent 3D scenes with unprecedented fidelity. 
Recent implicit representations are typically expressed as neural fields (NFs) and are created using differentiable volumetric rendering~\cite{mildenhall2021nerf}. 
This was first popularised by the neural radiance field (NeRF)~\cite{mildenhall2021nerf} and while it and follow-up works~\cite{martin2021nerf, barron2021mip, muller2022instant} have shown impressive visual rendering performance, their underlying surface geometry has a limited fidelity~\cite{wang2021neus} and are not smooth~\cite{oechsle2021unisurf}. 
Recently, NeuS~\cite{wang2021neus} has shown how volumetric rendering can be used to train signed distance functions (SDFs)~\cite{park2019deepsdf}, enabling neural fields with highly accurate geometric representations better suited for registration.

Neural field registration is important for their use in robotic applications, as it enables uses such as the fusion of multiple implicit maps, and the ability to dynamically update an existing implicit field. 
Nerf2nerf~\cite{goli2023nerf2nerf} was one of the first works to investigate neural field registration, by considering the registration problem as a deep learnt optimisation function between two neural fields. 
However, nerf2nerf relies on human-annotated keypoints for initialisation and assumes the scale of two neural fields are the same. 
In this paper, we propose \algoname{}, a novel method to estimate the relative 6-DoF pose transformation between two objects of interest which are located in two different neural fields. 
Our proposed method does not rely on human-annotated keypoints, operates directly on the continuous neural fields, and is capable of estimating transformation between two models with arbitrary scales. 
We build on nerf2nerf, proposing a bidirectional registration loss, the use of multi-view sampling of the NF surface, and the use of SDFs as the implicit model of choice.
These increase registration accuracy, take advantage of NFs ability to render data from any view, and ensure a consistent and clear geometric representation of implicit models respectively.

\begin{figure}[t]
    \centering
    \includegraphics[width=\linewidth, trim=0cm 0.5cm 0cm 0cm,clip]{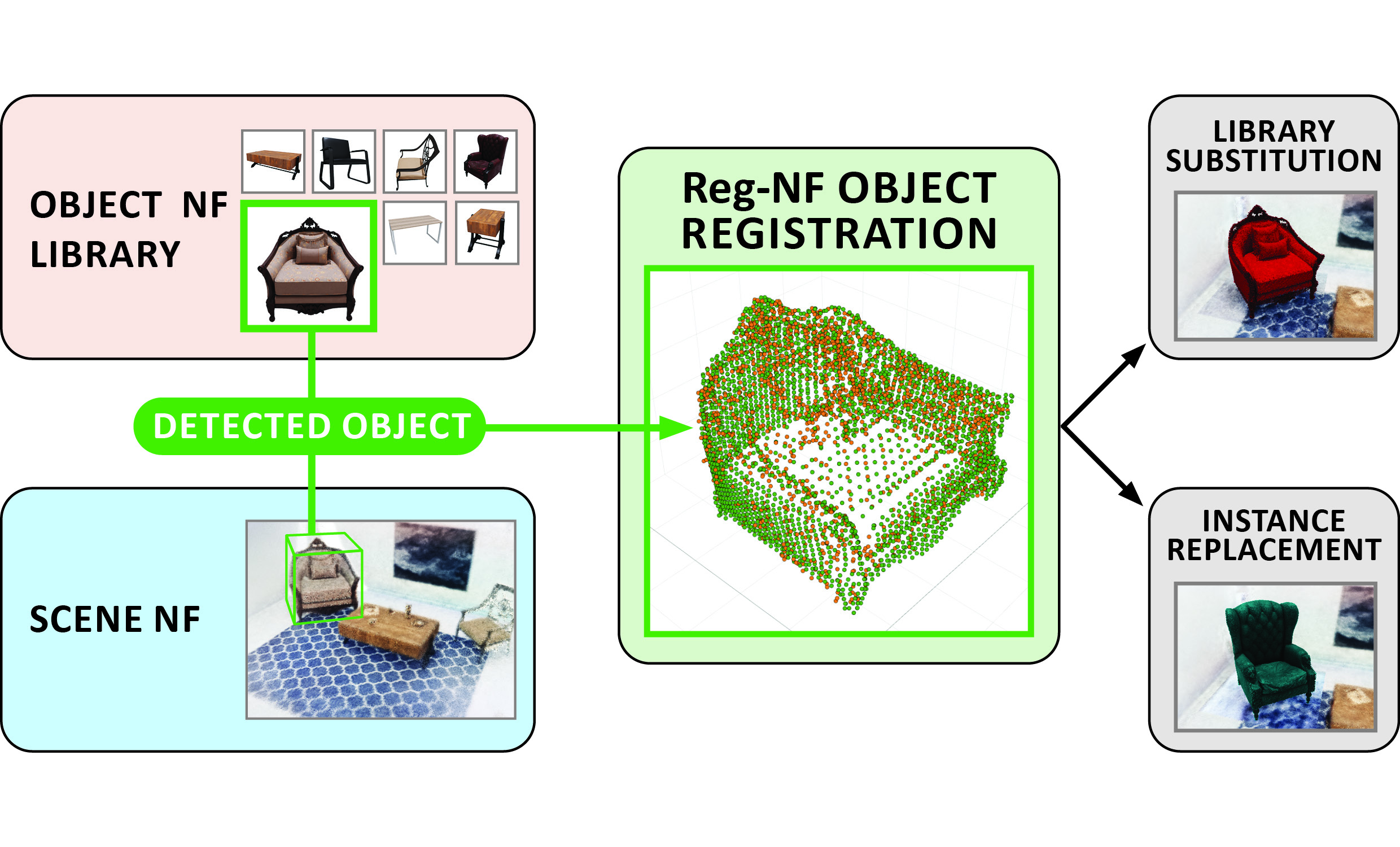}
    \vspace{-10mm}
    \caption{The proposed pipeline for using \algoname{} registration. Object is detected in a scene neural field (NF) and matched to an object in a library of object-centric NF. \algoname{} performs registration between the two NFs enabling neural substitution of the library object NF into the scene or the replacement of the object with another from the library. Substitution and replacement models coloured for clarity.}
    \label{fig:hero}
\vspace{-5mm}
\end{figure}

Our experiments highlight use-cases for NF registration where the 6DoF output from \algoname{} can be used for NF editing to merge NFs together.
We consider the scenario where \algoname{} is used to register a large scene NF with high-fidelity object-centric NFs stored in an object library, enabling both substitution of library objects into the scene, and replacement of object instances within the scene as shown in Fig.~\ref{fig:hero}.
This shows two particular benefits for \algoname{} in robotics. 
The first is object completion, using library substitution to improve the representation of scenes that have only partially observed objects or have under-trained NFs due to hardware constraints.
The second is using instance replacement as a way to enable data-driven simulation where any scene NF can be edited (by replacing objects instances) and used as new data for training in simulated NF environments. 

\section{Related Work}
\label{sec:related}

\subsection{Neural Fields} 

Neural fields (NFs) are implicit representations of 3D space that have gained great popularity in recent years~\cite{barron2021mip, mildenhall2021nerf, wang2021neus, martin2021nerf, yariv2021volume, Fu2022GeoNeus, kobayashi2022distilledfeaturefields, muller2022instant}. NFs use a small neural network to map any point in a normalised 3D space $\textbf{x} \in \mathbb{R}^3$ to the field's desired output/s (\eg{} colour, density, opacity, etc.). 

The recent popularity of neural fields is attributed to the introduction of Neural Radiance Fields (NeRFs)~\cite{mildenhall2021nerf} which has spawned many radiance field derivatives~\cite{barron2021mip, barron2022mip, kobayashi2022distilledfeaturefields}. 
The NeRF model is a neural field which maps $\textbf{x}$ and a viewing angle $\textbf{v} \in \mathbb{R}^2$ to view-dependant colour $\textbf{c}(\textbf{x}, \textbf{v}) \in \mathbb{R}^3$ and view-independent density $\sigma(\textbf{x}) \in \mathbb{R}$.
Density represents the likelihood of $\textbf{x}$ hitting anything in space.
With this model, any pixel can be represented as a ray ($\textbf{r}$)  described by origin ($\textbf{o}$) and normalised orientation ($\textbf{v}$), and the colour of that pixel can be computed through volumetric rendering~\cite{mildenhall2021nerf}. NeRF can be trained using only images and associated camera poses by comparing the true colour of a given pixel to the one rendered from the NeRF.

Despite it's strengths, NeRF's focus is on visual rendering which can lead to inaccurate geometric representations~\cite{yariv2021volume, wang2021neus}.
Surface neural network models $(S)$ ~\cite{park2019deepsdf,chibane2020ndf, wang2021neus, Fu2022GeoNeus, yariv2021volume}, seek to solve this by describing a continuous function $f$ mapping $\textbf{x}$ to the distance to the nearest surface $f(\textbf{x}) \in \mathbb{R}$ with the surface located wherever $f(\textbf{x}) = 0$.
These are known to provide smooth and consistent geometric representations~\cite{wang2021neus}. To enable training through purely image data as done by NeRF, some works combine the volumetric representation with implicit surface models~\cite{wang2021neus,Fu2022GeoNeus, yariv2021volume} and it is these models that will be the focus within our work, outlined in more detail within Section~\ref{sec:method}.

\subsection{Neural Fields for Robotics} 

Unlike explicit scene representations, such as voxels, point clouds, meshes, and surfels \cite{park2021elasticity}, NFs have attractive properties for robotics as their representations are continuous  and memory efficient. Recent work has extended the uses of NFs beyond synthesising novel views to robotics applications. 
Prior methods \cite{adamkiewicz2022vision,kwon2023renderable, maggio2023loc} demonstrate using the gradient of density from the NFs for collision-free robot motion planning. Several approaches \cite{yen2022nerf, li20223d, weng2023neural} show how continuous representation of NFs can be combined with additional constraints to jointly optimise for grasp and motion planning. 

Many algorithms have recently been developed to extend Neural Fields in simultaneous localisation and mapping (SLAM). iMAP\cite{sucar2021imap} is the first method to show NeRF-enabled SLAM with the aid of depth measurements from RGB-D sensors to reconstruct room-size scenes in real-time. NICE-SLAM\cite{zhu2022nice} introduces a hierarchical implicit representation to represent larger scenes, and NICER-SLAM\cite{zhu2023nicer} applies neural implicit representations for RGB-only SLAM and shows promising real-time properties.  

\subsection{Neural Fields Registration} 

Registration is the task of estimating the relative transformation between two 3D scene models. It has been extensively studied on explicit representations (\eg{} point clouds), whilst registering multiple implicit scene representations remains an underexamined challenge. %
Nerf2nerf \cite{goli2023nerf2nerf} is the first work which demonstrates relative transformation estimation between two NeRF models. However, they rely on human-annotated keypoints for initialisation, which is not practical in robotics applications. DReg-NeRF \cite{chen2023dreg} uses NeRF models then coverts them to an occupancy voxel grid to train 3D CNN followed by attention layers to learn the relations between the pairwise feature grids.  DReg-NeRF estimates transformation on explicit and discrete representation. Zero-NeRF\cite{peat2022zero} performs image to image registration leveraging NeRF representations but in the image space as opposed to between 3D scenes. 
Moreover, all the exiting prior works assume the scale of the two models are the same. 
In reality, each model's coordinate frame will be normalised to fit their specific training conditions. 
This does not guarantee consistent scale, particularly when considering models trained to represent individual objects and those trained to represent scenes.
By comparison, our proposed method does not rely on human-annotated keypoints, and is capable of estimating transformation between two models with arbitrary scales.

\section{Methodology}
\label{sec:method}
\subsection{Reg-NF Overview}

\begin{figure}[t]
    \centering
    \includegraphics[width=0.95\linewidth]{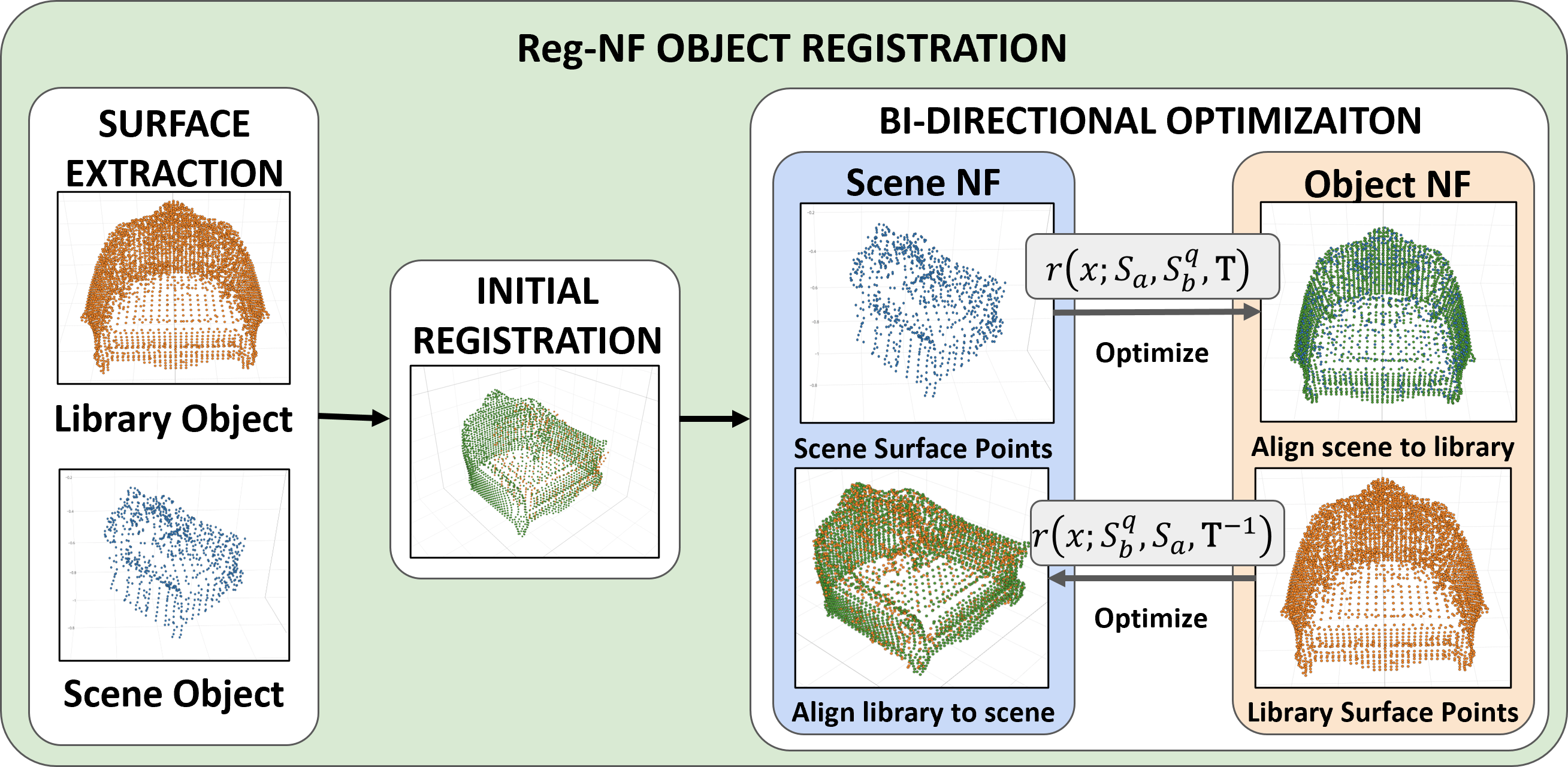}
    \vspace{-3mm}
    \caption{Overview of our \algoname{} registration process. Blue and orange denotes surface sample points from the scene and library NFs for a matched object respectively. Green points represent the target alignment during optimisation. After surface extraction and an initial registration estimate, bi-directional optimisation iterates till convergence. Final output is a 6-DoF transformation matrix between models.}
    \label{fig:method}
\vspace{-5mm}
\end{figure}

In Reg-NF, we provide a technique for aligning the surfaces of two different SDF NFs, by minimising the difference between their surfaces values. 
Minimisation is optimised for the 6-DoF pose transformation $\textbf{T}$ between the two NFs. We calculate $\textbf{T}$ using a differentiable optimisation function, initialised with an automated procedure. While Reg-NF is generalisable across different use-cases, in this paper we aim to find $\textbf{T}$ between a detected object in a larger scene NF with one in a pre-trained object NF library. We denote $a$ as the notation of an implicit representation of a scene, and $b^q$ as the $q$th object-specific implicit model from a database of object neural fields where $q \in \{1, ... , Q\}$.  
Please see  Fig.~\ref{fig:method} and Fig.~\ref{fig:hero} for an overview of Reg-NF and our considered use-case pipeline respectively.

\subsection{Preliminaries}

Our method assumes we have access to both scene and object NF models.
As background information, we provide a brief overview of the NF representation used for our work and its training process.

We utilise volumetric implicit surface fields, specifically NeuS~\cite{wang2021neus} for all NFs in our work. These provide colour $c(\textbf{x},\textbf{v})$ and signed surface distance $f(\textbf{x})$ mapping functions derived from a shared backbone network with separate SDF and colour output heads.

For any given pixel in an image, points at distance ($t$) along the ray emitted from this pixel are represented as $\textbf{r}(t) = \textbf{o} + t\textbf{v} | t \geq 0$.
Given a sampling of points along a ray in the range $[t_1, t_2, ..., t_n]$ of increasing magnitude, volumetric rendering can be used to calculate the colour of a given pixel:

\begin{equation}
    \hat{C}(\textbf{r}, \textbf{v}) = \sum_{i=1}^{n} T_i\phi_ic(\textbf{r}(t_i), \textbf{v}),
\end{equation}
where $T_i$ is the discrete accumulated transmittance at the $i$th point defined by $T_i = \prod_{j=1}^{i-1} (1 - \phi_j)$ and $\phi_i$ is the discrete opacity at the $i$th point along the ray.
Discrete opacity in NeuS is akin to density in NeRF only it is calculated based on the neural SDF output at the given point along the ray.

Training of the model is done by randomly sampling a batch of ($m$) pixels from a set of training images with known camera poses such that we get training data $D = \{ C(\textbf{r}_k), \textbf{o}_k, v_k\}$ from which $n$ points are sampled along each ray.
Loss is then computed as $L = L_c + \lambda L_e$ where $L_c$ is the colour loss term defined by $L_c = \frac{1}{m} \sum_k |\hat{C}(\textbf{r}_k) - C(\textbf{r}_k)|$, $L_e$ is an Eikonal loss regularisation term defined as $L_e = \frac{1}{nm} \sum_{k,i} (||f'(\textbf{r}_k(t_i))||_2 - 1)^2$, and $\lambda$ is a weighting factor for the Eikonal regularisation. For more details, refer to~\cite{wang2021neus}.

\subsection{Automated initialisation}

\subsubsection{Initial multi-view surface sampling}

\algoname{} begins by establishing approximate correspondences between objects of interest within $a$ and $b^q$. 
We assume an object detection provides approximate location and classification of an object in $a$ that matches to $b^q$. 
Based on this, we calculate a set of $N$ camera locations (extrinsics, $E_n$), which are all oriented to view the centroid of the detected object of interest. 
We generate a grid pattern of rays travelling from each camera pose $E_n$ and sample points along each ray.
For each ray we return surface sample points at the first point along the ray where $f(\textbf{x})= 0$.
These initial surface sample points for each of our $N$ camera poses are then merged, providing object surface sample point sets $P_a$ and $P_b^q$, for each object in our two respective neural fields.
This multi-view sampling approach takes advantage of the ability of NFs to render data from any view, providing a clear geometric representation of the object for initialisation.

\subsubsection{Calculate an initial transformation}

To find the initial estimate of the transformation, we employ RANSAC~\cite{fischler1981random} with Fast Point Feature Histogram (FPFH)~\cite{rusu2009fast} descriptors to estimate the correspondence between source and target. RANSAC provides an initial alignment approximation which is further refined via the point-to-point Iterative Closest Point (ICP)~\cite{121791} method. 
Through this, for two initial sets of surface points $P_a$ and $P_b^q$, we attain the initial six-degree-of-freedom (6-DoF) pose transformation $\hat{\textbf{T}}$.

\subsection{Bidirectional Registration Loss}

We establish neural field registration as an optimisation function that uses gradient decent to find the optimal pose transformation $\textbf{T}$ between two surface fields. The parameters of our optimisation are initialised using $\hat{\textbf{T}}$, and assuming an initial scale factor of $s = 1$. Our neural field registration problem can be expressed by the following equation:
\begin{equation}
    \argmin_\textbf{T} L_s \left(S_a, S_b^q; \textbf{T} \right),
\end{equation}
where $S_a$ and $S_b^q$ denotes the signed distance representation for implicit models $a$ and $b^q$. 

In nerf2nerf~\cite{goli2023nerf2nerf}, optimisation is performed over a discrete set of samples collected from a single implicit model, with a robust kernel $\kappa$ used to improve the robustness against outlier samples. The kernel contains learnt parameters $p$ and $\alpha$ which control the decision boundary and impact of outlier samples on registration. In our approach, we consider that the accuracy of optimisation can be improved by collecting samples from both $S_a$ and $S_b^q$ ($A$ and $B^q$ respectively), and performing a bidirectional optimisation over these surfaces. Furthermore, we include a regulariser, which is designed to penalise the function when $A$ and $B^q$ are deviating from each other.
\begin{equation}
\begin{split}
    L_s(S_a, S_b^q; \textbf{T}) = E_{x \in A} \kappa (r(x; S_a, S_b^q, \textbf{T}); p, \alpha) \\
    + E_{x \in B^q} \kappa (r(x; S_b^q, S_a, \textbf{T}^{-1}); p, \alpha) 
    + w L_r ,
\end{split}
\end{equation}
where $L_r$ is our regulariser with a weight factor $w$. Our loss function $L_s$ is calculated using the current estimated pose transform $\textbf{T}$, which we compose by multiplying three transformation matricies together, representing rotation, translation and scale factor components:
\begin{equation}
    \textbf{T} = T \cdot R \cdot \sigma , 
\end{equation}
where $T$ denotes a 6-DoF translation transformation matrix with components $[t_x, t_y, t_z]$, $R$ denotes a rotation transformation matrix composed from Euler angles $[r_r,r_p,r_y]$, and $\sigma$ denotes a scale transformation matrix composed by scale factor $s$. 
Note that we use a scalar scale factor across all axes to avoid object warping. 
Our loss function is optimised over the learnt parameters: $(t_x,t_y,t_z,r_r,r_p,r_y,s,p,\alpha)$.

\subsection{Registration Residuals and Sampling Procedure}
Our registration residuals $r$ are expressed as:
\begin{equation}
    r(x; S_a, S_b^q, \textbf{T}) = \| S_a - S_b^q\textbf{T} \| \quad x \in A ,
\end{equation}
and in the bidirectional case:
\begin{equation}
    r(x; S_b^q, S_a, \textbf{T}^{-1}) = \| S_b^q - S_a\textbf{T}^{-1} \| \quad x \in B^q .
\end{equation}
Therefore, the residuals are calculating the difference in surface values between SDFs $S_a$ and $S_b^q$. It is important to note that since SDF values are continuous, this loss formulation is differentiable. As SDF values increase in absolute magnitude the further they are away from a surface, the gradient descent will optimise the transformation $\textbf{T}$ to match the surfaces even if the initial samples $A$ and $B^q$ have poor initialisation.

Initially, we set samples $A$ and $B^q$ equal to the initial surface samples $P_a$ and $P_b^q$. We allow for new samples to be discovered using a variant of the Metropolis-Hastings sampling procedure in nerf2nerf~\cite{goli2023nerf2nerf}. Our updated sampling algorithm is shown in Algorithm 1. Because SDF values are continuous and not discrete, the sampling strategy needs to be carefully designed to prevent the generation of samples away from true object surfaces.

\subsection{Regularisation Loss}
Our regularisation loss ($L_r$) term calculates the mean distance in 3D sample positions between the two different sets of sample points (one from each direction of optimisation); ${A, B^q}$. 
For point correspondence, we assume the nearest 3D point in the other set of sample points ${B^q}$ is the ground-truth corresponding point given a particular sample point from ${A}$ (similar to ICP~\cite{121791}).
\begin{equation}
    L_r = \sum \frac{\min_{B^q} D_{A,B^q}^2}{\vert A \vert}, D_{A,B^q}^2 = \| A - B^q\textbf{T} \|^2,
\end{equation}
where $\vert A \vert$ denotes the number of samples in $A$. Given some sample points $A$ from $S_a$, and a different set of transformed (by $\textbf{T}$) sample points $B^q$ from $S_b^q$, after optimisation, their 3D points should be approximately the same. 
The bidirectional components are balanced by $L_r$, which also acts to penalise `unmatched points' between the two models. 
For example, if one set of samples covers the entirety of a large table in one SDF, and the second set of samples only considers a small proportion of the same table object in the second SDF, then a large disparity will be identified by the regulariser.
\setlength{\textfloatsep}{1pt}
\begin{algorithm}[t]
\caption{\algoname{} Sampling Procedure from $S_a$}\label{alg:cap}
$C \leftarrow A^{(t-1)} + \rho \cdot U_3[-1,+1]$\;
\For{ \text{each sample} $x \in C$}
{
 
 \If{$S_a(x) \leq \omega_1$ \Algand $S_b^q(x\textbf{T}) \leq \omega_2$
 \Algand $r(x; S_a, S_b^q, \textbf{T}) \leq \xi$ \Algand $d(x, A^{(t-1)}) \geq \rho/10$}
 {
    $A_{(t)} \leftarrow A_{(t)} \cup \{x\} $ \;
    
  }
}
\end{algorithm}

\section{Experimental Design}

\begin{figure}
    \centering
    \subfloat[ac-room]{\includegraphics[width=0.25\linewidth]{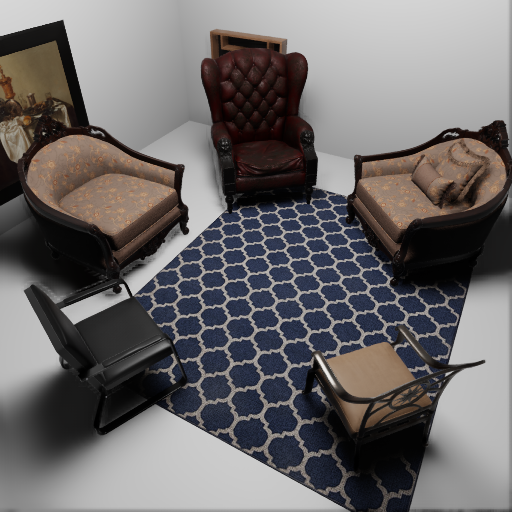}}
    \quad
    \subfloat[atbl-room]{\includegraphics[width=0.25\linewidth]{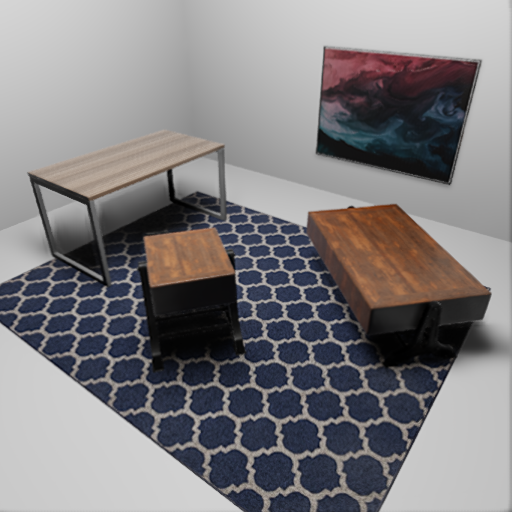}}
    \quad
    \subfloat[mix-room]{\includegraphics[width=0.25\linewidth]{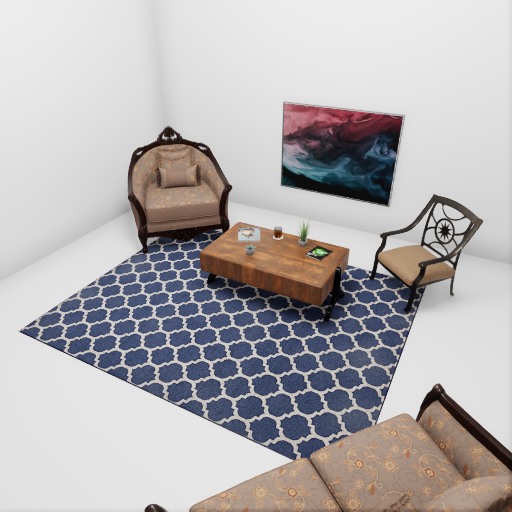}}
    \caption{Overview of scenes and object models in \dataname dataset. Chairs objects shown in ac-room, starting clockwise from bottom-left are: chair (\textit{c}), fancy chair with no pillow (\textit{fc-nop}), matrix chair (\textit{mc}), dining chair (\textit{dc}), and fancy chair (\textit{fc}). Tables shown in atbl-room from left to right are: willow table (wt), end table (et), and table (t). Objects of interest in mix-room are fc, t, and dc.}
    \label{fig:datasets}
\end{figure}

\noindent\textbf{Dataset}: The dataset we use for our experiment comprises high-fidelity simulated images and corresponding camera poses of objects and scenes, collected using NVIDIA's Omniverse Isaac Sim platform.
We will refer to this dataset as our object NF registration (ONR) dataset.
Object data is of single objects in a ``void'' and scene data is a standardised room with different object models placed within.
Object data becomes the basis for our object NF library and contains data for 5 chair and 3 table models learnt at varying scales to maximize fidelity.
Data for scenes is collected with manually generated trajectories under two conditions: long and short trajectories.
Long trajectories maximise coverage of all objects within the scene while short ones do a more general pass of the room.
These represent best-case and more realistic data capture for scene NF training respectively.
There are three scenes collected: 1) containing all chair models stored as object data (ac-room); 2) containing all table models stored as object data (atbl-room); 3) a scene with a mixture of two chairs and a cluttered table (mix-room). 
Example images from each scene and individual object names are shown in Fig.~\ref{fig:datasets}.

\ifconffinal
\begin{table*}[t]
\centering
\caption{Comparison between \algoname{} and nerf2nerf on \dataname Dataset.}
\label{tbl:n2n_compare}
\begin{tabular}{cc|ccccc|ccc|ccc}
 &               & \multicolumn{5}{c|}{\textbf{All Chairs Room}}                                      & \multicolumn{3}{c|}{\textbf{All Tables Room}} & \multicolumn{3}{c}{\textbf{Mix Room}}             \\ \cline{3-13} 
  &               & c              & dc             & fc             & fc-nop         & mc             & et            & t            & wt             & dc             & fc             & t              \\ \hline
\multicolumn{1}{c|}{\multirow{2}{*}{\textbf{$ \Delta \mathbf{t}\downarrow$}}}                   & nerf2nerf~\cite{goli2023nerf2nerf}           & 0.278          & 0.291          & 0.127          & 0.117          & \textbf{0.0007}          & 0.525              & 0.084            & 0.106               & 0.343          & 0.154          & 0.086          \\
\multicolumn{1}{c|}{}                                                         & \textbf{\algoname} (ours) & \textbf{0.04} & \textbf{0.035} & \textbf{0.018} & \textbf{0.014} & 0.007 & \textbf{0.398}           & \textbf{0.044}          & \textbf{0.022}          & \textbf{0.292} & \textbf{0.029} & \textbf{0.009} \\ \hline
\multicolumn{1}{c|}{\multirow{2}{*}{\textbf{$ \Delta \mathbf{R}\downarrow$}}} & nerf2nerf~\cite{goli2023nerf2nerf}           & \textbf{0.041}          & 0.088          & 0.050          & 0.034 & \textbf{0.002} & \textbf{2.396}              & \textbf{0.002}             & 0.221               & \textbf{0.050} & 0.034          & 0.026          \\
\multicolumn{1}{c|}{}                                                         & \textbf{\algoname} (ours) & 0.048 & \textbf{0.039} & \textbf{0.031} & \textbf{0.020}          & 0.009          & 2.6           &    0.053         & \textbf{0.025}          & 0.641          & \textbf{0.030} & \textbf{0.012} \\ \hline
\multicolumn{1}{c|}{\multirow{2}{*}{\textbf{$ \Delta s \downarrow$}}} &    nerf2nerf~\cite{goli2023nerf2nerf}       & NA       & NA       & NA       & NA        & NA    & NA              & NA              & NA           & NA          & NA          & NA          \\
\multicolumn{1}{c|}{} &
\textbf{\algoname} (ours)     &    \textbf{0.019}     &  \textbf{0.007}      &  \textbf{0.007}      &    \textbf{0.060}      &     \textbf{0.003}  &      \textbf{0.019}           &       \textbf{0.004}          &       \textbf{0.006}      &      \textbf{0.021}       &        \textbf{0.009}     &      \textbf{0.005}       \\
 \hline
\end{tabular}
\end{table*}
\fi

\ifarxiv
\begin{table*}[t]
\centering
\caption{Comparison between \algoname{} and nerf2nerf on \dataname Dataset.}
\label{tbl:n2n_compare}
\begin{tabular}{cc|ccccc|ccc|ccc}
 &               & \multicolumn{5}{c|}{\textbf{All Chairs Room}}                                      & \multicolumn{3}{c|}{\textbf{All Tables Room}} & \multicolumn{3}{c}{\textbf{Mix Room}}             \\ \cline{3-13} 
  &               & c              & dc             & fc             & fc-nop         & mc             & et            & t            & wt             & dc             & fc             & t              \\ \hline
\multicolumn{1}{c|}{\multirow{3}{*}{\textbf{$ \Delta \mathbf{t}\downarrow$}}}                   & FGR~\cite{zhou2016fast} & 0.868 & 1.084 & 0.3715 & 0.485 & 0.152 & 0.368 & 0.819 & 0.219 & 1.351 & 0.226 & 0.636\\
\multicolumn{1}{c|}{}  & nerf2nerf~\cite{goli2023nerf2nerf}           & 0.278          & 0.291          & 0.127          & 0.117          & \textbf{0.0007}          & 0.525              & 0.084            & 0.106               & 0.343          & 0.154          & 0.086          \\
\multicolumn{1}{c|}{}                                                         & \textbf{\algoname} (ours) & \textbf{0.04} & \textbf{0.035} & \textbf{0.018} & \textbf{0.014} & 0.007 & \textbf{0.398}           & \textbf{0.044}          & \textbf{0.022}          & \textbf{0.292} & \textbf{0.029} & \textbf{0.009} \\ \hline
\multicolumn{1}{c|}{\multirow{3}{*}{\textbf{$ \Delta \mathbf{R}\downarrow$}}} & FGR~\cite{zhou2016fast} & 1.130 & 1.669 & 1.020 & 1.449 & 0.1389 & 2.513 & 1.656 & 1.282 & 1.669 & 0.395 & 2.297\\
\multicolumn{1}{c|}{}  & nerf2nerf~\cite{goli2023nerf2nerf}           & \textbf{0.041}          & 0.088          & 0.050          & 0.034 & \textbf{0.002} & \textbf{2.396}              & \textbf{0.002}             & 0.221               & \textbf{0.050} & 0.034          & 0.026          \\
\multicolumn{1}{c|}{}                                                         & \textbf{\algoname} (ours) & 0.048 & \textbf{0.039} & \textbf{0.031} & \textbf{0.020}          & 0.009          & 2.6           &    0.053         & \textbf{0.025}          & 0.641          & \textbf{0.030} & \textbf{0.012} \\ \hline
\multicolumn{1}{c|}{\multirow{3}{*}{\textbf{$ \Delta s \downarrow$}}} &    FGR~\cite{zhou2016fast} & NA & NA & NA & NA & NA & NA & NA & NA & NA & NA & NA\\
\multicolumn{1}{c|}{}  & nerf2nerf~\cite{goli2023nerf2nerf}       & NA       & NA       & NA       & NA        & NA    & NA              & NA              & NA           & NA          & NA          & NA          \\
\multicolumn{1}{c|}{} &
\textbf{\algoname} (ours)     &    \textbf{0.019}     &  \textbf{0.007}      &  \textbf{0.007}      &    \textbf{0.060}      &     \textbf{0.003}  &      \textbf{0.019}           &       \textbf{0.004}          &       \textbf{0.006}      &      \textbf{0.021}       &        \textbf{0.009}     &      \textbf{0.005}       \\
 \hline
\end{tabular}
\vspace{-6mm}
\end{table*}
\fi

\noindent\textbf{Metrics}: We follow the metrics in nerf2nerf~\cite{goli2023nerf2nerf} for our work.
We report the root mean squared error (RMSE) between the ground-truth and predicted scene to object transformation matrices.
Rotation error $\Delta \mathbf{R}$ is calculated in radians and translation error $\Delta \mathbf{t}$ is calculated in normalised object frame units.
We also report the absolute difference between estimated and true scale between scene and object models $\Delta s$.
Note that this is given instead of RMSE as scale factor is assumed consistent across all axes.

\noindent\textbf{Nerf2nerf on ONR dataset}: To evaluate nerf2nerf on the ONR dataset, we enabled nerf2nerf to utilise the same surface fields from our SDF models as used by \algoname{}. The only modification to nerf2nerf was to modify their sampling algorithm to suit SDF notation, where a surface has a value of zero. We manually generated new human annotated keypoints for the initialisation procedure used in nerf2nerf. 

\noindent\textbf{Training models}: All NF models were trained using the sdfstudio~\cite{Yu2022SDFStudio} implementation of NeuS which includes the proposal network from MipNeRF-360~\cite{barron2022mip} for training speed-up (neus-facto). For more details please refer to sdfstudio~\cite{Yu2022SDFStudio}.
Object models were trained for 30,000 iterations and scene models were trained for 100,000 iterations unless stated otherwise for specific experiments.

\noindent\textbf{Object Proposals}: \algoname{} assumes a detection has already been made within a scene's NF through some pre-existing method.
As object proposal generation is not within the scope of this work, we utilise ground-truth object 3D bounding boxes to calculate initial set of $N$ camera extrinsics for generating the initial surface samples. We also remove any samples that are far away from the object's 3D location.

\noindent\textbf{\algoname{} hyperparameters}: We provide the following hyperparameters for Reg-NF. For our sampler, we use $\omega_1=0.01$, $\omega_2 = 0.02$ and $\xi = 0.02$. We set $\rho$ to $r/20$, where $r$ is the scene radii and generate new samples every $10$ iterations. We use a learning rate of $0.02$ for rotation, $0.01$ for translation, $0.01$ for scale, and $0.005$ for adaptive kernel parameters, for a maximum of $200$ iterations. We also have early stopping criteria, when $\sum (r(x; S_a, S_b^q, \textbf{T}), \, x \forall A)/\vert A \vert \leq 0.0005$.

\section{Results}
We first perform a quantitative analysis of \algoname{}, comparing it to nerf2nerf and demonstrate that we are outperforming them while not requiring manually annotated keypoints or an assumption that all objects are of the same scale as the scene.
This is followed by two experiments demonstrating the robustness of \algoname{} to scale and the benefit of \algoname{} multi-view surface extraction. Finally, we demonstrate the benefits of \algoname{} for modelling imperfect scene NFs with known object NF replacement, and show how \algoname{} can enable object instance replacement for modelling alternative NF scenes with the same underlying object arrangements but different object NF models.

\subsection{Comparison to nerf2nerf}

We evaluate and compare the performance of \algoname{} and nerf2nerf~\cite{goli2023nerf2nerf}, on our \dataname dataset. We also compare against FGR for completeness~\cite{zhou2016fast}.
Results shown in Table~\ref{tbl:n2n_compare} are the average results across 10 iterations of the experiment to account for randomized factors such as RANSAC.
In Table~\ref{tbl:n2n_compare} we see that \algoname{} is typically at least an order of magnitude better than nerf2nerf in terms of $\Delta \textbf{t}$ and is still generally superior in $\Delta \textbf{R}$. We attribute the large increase of errors for nerf2nerf as being primarily due to the inherent scale differences between scene and database object models, for which nerf2nerf has no functionality to handle. 

\begin{figure}
    \centering
    \subfloat{\includegraphics[width=0.25\linewidth]{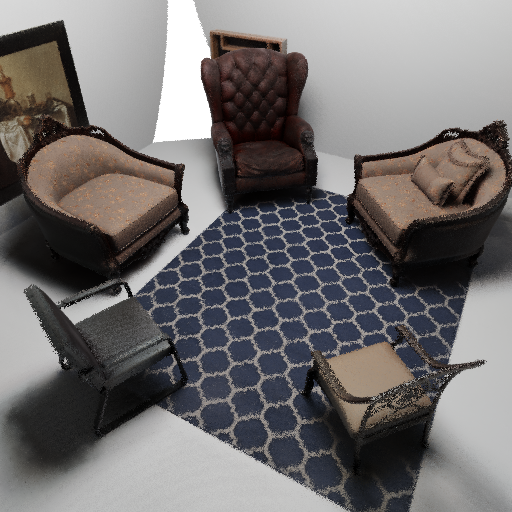}}
    \quad
    \subfloat{\includegraphics[width=0.25\linewidth]{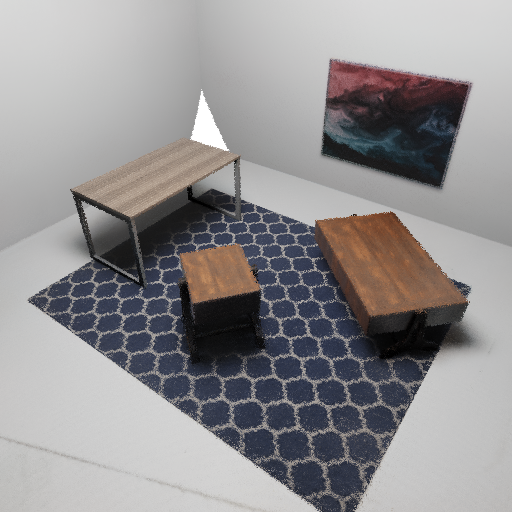}}
    \quad
    \subfloat{\includegraphics[width=0.25\linewidth]{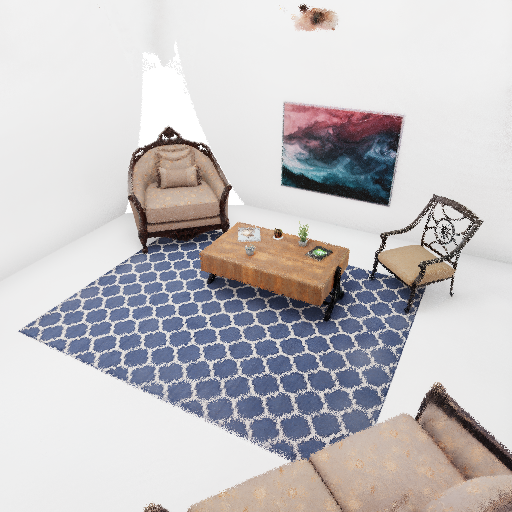}}
    
    \subfloat{\includegraphics[width=0.25\linewidth]{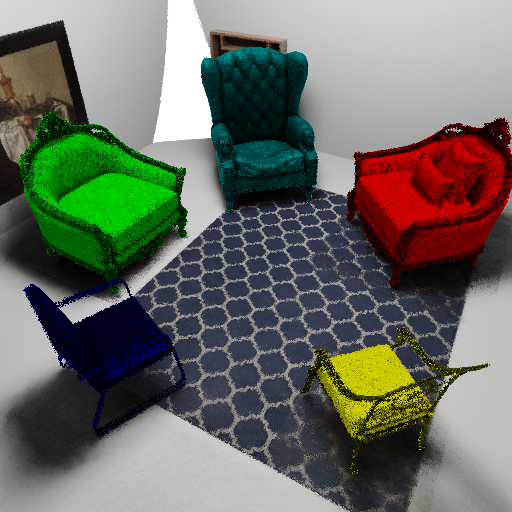}}
    \quad
    \subfloat{\includegraphics[width=0.25\linewidth]{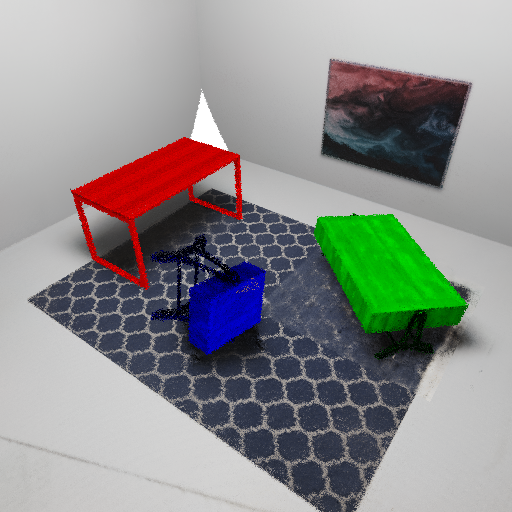}}
    \quad
    \subfloat{\includegraphics[width=0.25\linewidth]{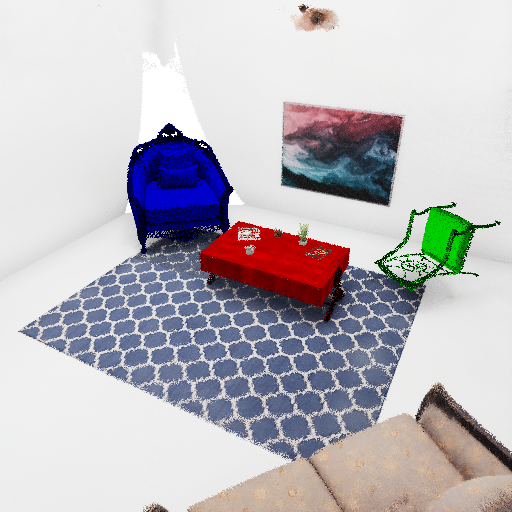}}
    \caption{Example of library replacement using \algoname{} for all objects in all scenes evaluated in Table~\ref{tbl:n2n_compare}. Top: Original scene NF render. Bottom: Scene NF with library object NF substitutions render. Substitutions are based on \algoname{} outputs. Note, colours are added to object NFs during render to provide visual distinction between scene NF and object NFs.}
    \label{fig:all-scene-replace}
\end{figure}

Focusing on \algoname{}, we note that failures can still occur, such as when we match object \emph{dc} to scene \emph{Mix Room} or \emph{et} to scene \emph{at-room}.
In both these cases, we note the cause of failure being poor initialisation that proved inescapable for \algoname{}.
A qualitative analysis of the \algoname{} registration can be seen in Fig~\ref{fig:all-scene-replace} where coloured versions of NF object library models are substituted into the original scene NF at their poses calculated by \algoname{}. 

\subsection{Robustness to scale}
We conducted an ablation study to investigate the performance of our approach under extreme scale differences between the two neural fields. 
We use Mix Room for our scene neural field and we attempt to align the \emph{fc} object within that room to library \emph{fc} models. 
For this test, we generated four additional \emph{fc} models, each resized with scale factors $2, 0.5, 0.25, 0.1$. 
For these experiments, as convergence takes longer with large scale differences, we ran our optimisation procedure for 1000 iterations. 
Our results are shown in Table~\ref{tbl:scale}. We observe that our approach can successfully align the two object fields at a scale factor of 0.5. 
Interestingly, our learnt scale optimisation can successfully approximate the true scale factor between the scene and library models, even at a scale factor of $2$. In extreme scale scenarios, we observe that the method has room for improvement.  

\begin{table}[]
\centering
\caption{\algoname{} scale analysis study on Mix Room.}
\label{tbl:scale}
\begin{tabular}{c|cccc}
 Scale Factor  & \textbf{$\Delta \mathbf{t}\downarrow$}  & \textbf{$ \Delta \mathbf{R}\downarrow$}             & Est. scale            & True scale (GT)       \\ \hline
   2 & 1.40 & 1.40 & 2.39 & 2.5 \\
   1 & 0.03 & 0.03 & 1.24 & 1.25 \\
   0.5 & 0.001 & 0.002 & 0.62 & 0.62 \\
   0.25 & 0.30 & 1.42 & 0.62 & 0.31 \\
   0.1 & 0.17 & 2.58 & 0.34 & 0.13

\end{tabular}
\end{table}
\vspace{-1mm}

\subsection{Effect of multi-view sample initialisation}
The benefit of multi-view sampling during initialisation is most felt when an object has no distinguishing characteristics or is only partially seen from a single viewing angle taken from the training data.
Using a single view this way introduces a high level of variability in object coverage.
To show the impact of this, we perform experiments on the ac-room scene, comparing multi-view sampling to using a single challenging view extracted from the training data that observes the object. 
These views never see the front of chairs and often view only part of the chair, representing worst-case scenarios.
The rest of the Reg-NF pipeline is kept consistent and quantitative results are shown in Table~\ref{tbl:single-vs-multi}.
It is shown that a bad viewpoint with poor samples drastically reduces performance as no meaningful features could be extracted to enable effective registration.

\begin{table}[]
\caption{Challenging Single view vs Multi-view Tests}
\label{tbl:single-vs-multi}
\centering
\begin{tabular}{cc|ccccc|}
\multicolumn{2}{c|}{}                                                                             & \textbf{c} & \textbf{dc} & \textbf{fc} & \textbf{fc-nop} & \textbf{mc}\\ \hline
\multirow{2}{*}{\textbf{$\Delta \mathbf{t}\downarrow$}} & \textbf{Single}        & 0.307      & 0.674       & 0.378       & 0.681           & 0.147   \\
                                                                         & \textbf{Multi. (ours)} & \textbf{0.040}      & \textbf{0.035}       & \textbf{0.018}       & \textbf{0.014}           & \textbf{0.007}       \\ \hline
\multirow{2}{*}{\textbf{$ \Delta \mathbf{R}\downarrow$}}                 & \textbf{Single}        & 0.931      & 2.201       & 2.182       & 0.842           & 0.144    \\
                                                                         & \textbf{Multi. (ours)} & \textbf{0.048}      & \textbf{0.039}       & \textbf{0.031}       & \textbf{0.020}      & \textbf{0.009}       \\ \hline
\multirow{2}{*}{\textbf{$ \Delta s \downarrow$}} & \textbf{Single}        & 0.044      & 0.047       & 0.498       & 0.339           & 0.178  \\
                                                                         & \textbf{Multi. (ours)} & \textbf{0.019}      & \textbf{0.007}       & \textbf{0.007}       & \textbf{0.060}           & \textbf{0.003}       \\ \hline
\end{tabular}
\end{table}

\subsection{Substitution within imperfect scene models}
\begin{figure}[t]
    \centering
    \subfloat[Original scene]{\includegraphics[width=0.35\linewidth]{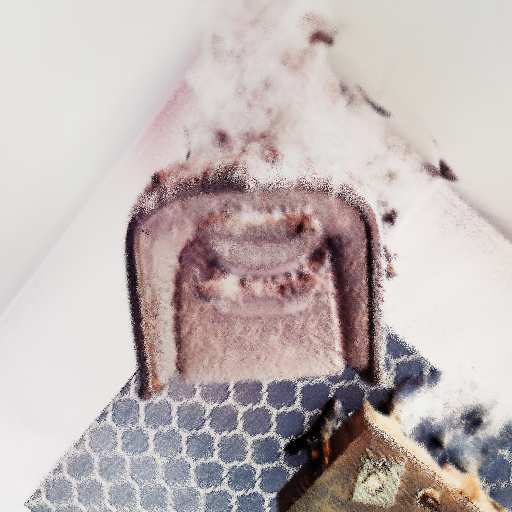}}
    \quad
    \subfloat[Library substitution]{\includegraphics[width=0.35\linewidth]{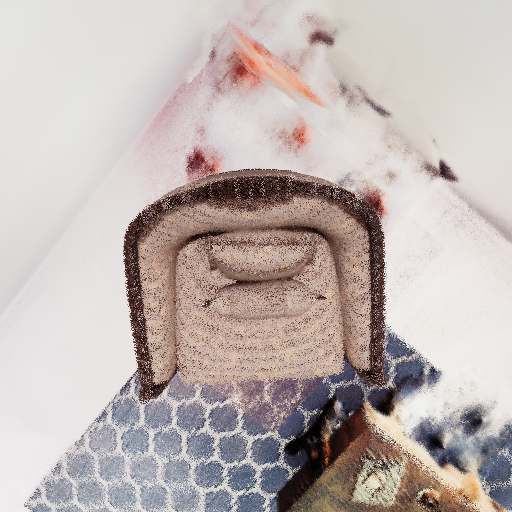}}
    \caption{Example of object completion via library replacement on a scene with low coverage. Original NF (a) is unable to correctly render the back of the object as it was not seen during training. (b) shows impact of library substitution from \algoname{} registration. Geometry of the object within the scene can be fully rendered from only partial initial view.}
    \label{fig:use-case-short}
\end{figure}

We examine two types of ``imperfect scene" to demonstrate practical applications for \algoname{}.
The first considers when a robot may not be able to fully traverse a scene to get ``full coverage'' of an object for the scene's NF.
For this we use the ``short" trajectories of our mix-room scene.
We can see in Fig.~\ref{fig:use-case-short} that the scene NF trained from the short trajectory is not able to render the back of the chair clearly as it has no information about that model.
Using \algoname{}, we register the object NF for the chair within the scene and can render a clear view of the back of the chair, fully understanding its geometry.

\begin{figure}[t]
    \centering
    \subfloat[Under-trained scene]{\includegraphics[width=0.35\linewidth]{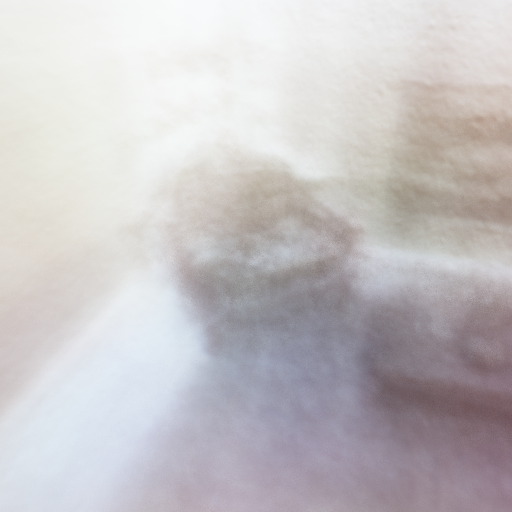}}
    \quad
    \subfloat[Library substitution]{\includegraphics[width=0.35\linewidth]{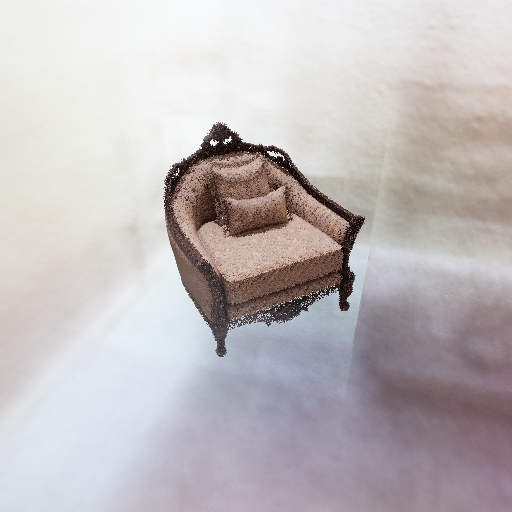}}
    \caption{Example of library replacement of object within under-trained scene NF. Original NF (a) has been under-trained but still needs to be used. (b) shows that if objects are detected, Reg-NF can provides needed transforms to replace the region of the detected object with a fully trained library model, providing clean object geometry.}
    \label{fig:use-case-undertrain}
\end{figure}

The second application setting considered for \algoname{} is operations with an under-trained scene model.
While there are techniques available for speeding up NF training~\cite{muller2022instant}, in robotics, hardware limitations may lead to situations where a scene model cannot be fully trained before needing to be used.
We therefore investigate the applicability of \algoname{} on highly under-trained scene NFs, trained with 300 steps rather than the 100,000 used in the main experiments.
While being an extreme example, we show in Fig.~\ref{fig:use-case-undertrain} that an accurate alignment and object NF substitution is still feasible for some objects in this setting, enabling a clear render of the object even while the rest of the scene is under-defined.

\subsection{Instance replacement}

\begin{figure}[t]
    \centering
    \subfloat[Original scene]{\includegraphics[width=0.35\linewidth]{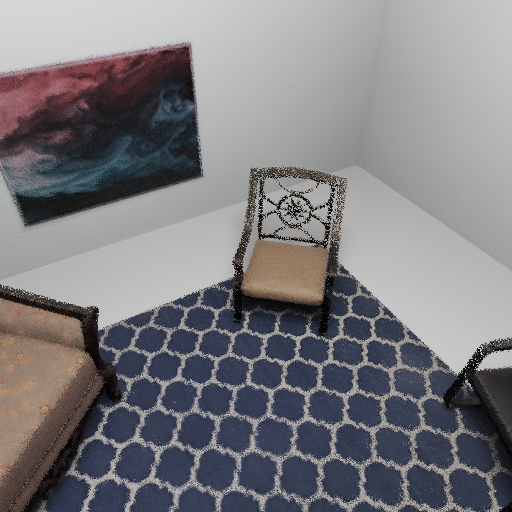}}
    \quad
    \subfloat[Instance replacement]{\includegraphics[width=0.35\linewidth]{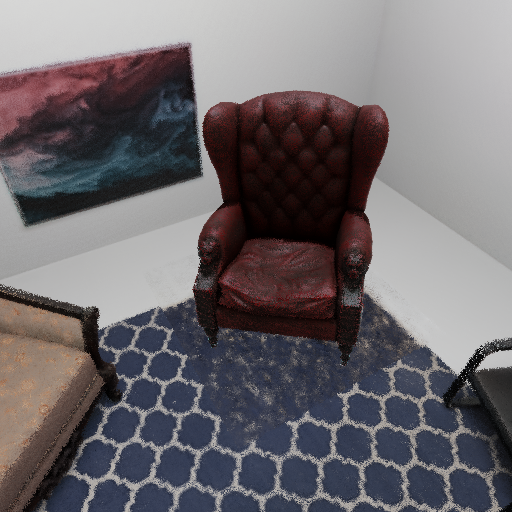}}
    \caption{Example of object instance replacement. After original chair in scene NF (a) has been properly registered against a matching object library NF instance via \algoname{}, it can be replaced by a different instance from the NF library (b).}
    \label{fig:instance replacement}
\end{figure}

Finally, we demonstrate the benefits of using a library of pre-trained NF objects for creating new scenes.
As all objects in the library are pre-defined and standardised, once \algoname{} derives the transform between a matched object NF and the scene NF, the known relative shapes/poses of objects within the NF library can be used to replace registered scene objects with any other object instance.
We demonstrate this within Fig.~\ref{fig:instance replacement}, showing \emph{dc} being replaced with \emph{mc}.
This opens up potential use-cases in data-driven robotics where, after a scene has been turned into a neural field, objects within that scene can be changed to provide new data based on the layout of the original scene.

\subsection{Limitations}
The primary limitation of \algoname{} is the initialisation process.
This is consistently the most extreme failure case in our work, when an initialisation is so poor that \algoname{} cannot find an optimal solution.
This is a problem not faced by nerf2nerf due to the manual keypoint selection. We also acknowledge that current results have assumed a direct mapping from a scene object to a known object model of the same type, an assumption that cannot hold true for all real-world settings.
This opens avenues of future work on template warping from a known library of objects.

\section{Conclusions}
We present \algoname{}, a novel method for registration between neural field (NF) representations.
Specifically, we estimate the 6DoF transform between objects found in a scene NF and object-centric NF counterparts stored in an NF object library, even when objects and scene have different scaling factors.
We introduce a bi-directional registration loss and utilise the continuous nature of NF representations to align surfaces between objects and the scene.
We analyse the effectiveness of \algoname{} and show its advantages for modelling objects within imperfect scene NFs and for enabling data-driven robotics research by offering editable scene NFs for robots to train in.

\balance{}

\bibliographystyle{IEEEtran}
\bibliography{refs}

\end{document}